\documentstyle[nemlap3]{article}

\input{epsf}    % include EPSF figures
\input{ipamacs} % include the WSUIPA font (better than nothing)

\title{Modularity in Inductively-Learned Word Pronunciation Systems
\thanks{This research was partially performed by the first and second author
at the Department of Computer Science of the Universiteit Maastricht
(The Netherlands), and partially in the context of the ``Induction of
Linguistic Knowledge'' research programme, partially supported by the
Foundation for Language Speech and Logic (TSL), funded by the
Netherlands Organization for Scientific Research (NWO).}}
\author{Antal van den Bosch$^{1}$,\ \ \  Ton Weijters$^{2}$,\ \ \  Walter
Daelemans$^{1}$ \ \\[0.25cm]
\begin{tabular}{ccc}
$^{1}$ ILK / Computational Linguistics & & $^{2}$ Department of Information
Technology \\
Tilburg University & & Eindhoven University of Technology \\
P.O. Box 90153 & & P.O. Box 513 \\
NL-5000 LE Tilburg & & NL-5600 MB Eindhoven \\
The Netherlands & & The Netherlands \\
{\tt \{antalb,walter\}@kub.nl} & & {\tt A.J.M.M.Weijters@tm.tue.nl} \\
\end{tabular}}

\begin{document}
\maketitle

\begin{abstract}
In leading morpho-phonological theories and state-of-the-art
text-to-speech systems it is assumed that word pronunciation cannot be
learned or performed without in-between analyses at several
abstraction levels (e.g., morphological, graphemic, phonemic,
syllabic, and stress levels). We challenge this assumption for the
case of English word pronunciation. Using {\sc igtree}, an
inductive-learning decision-tree algorithms, we train and test three
word-pronunciation systems in which the number of abstraction levels
(implemented as sequenced modules) is reduced from five, via three, to
one. The latter system, classifying letter strings directly as mapping
to phonemes with stress markers, yields significantly better
generalisation accuracies than the two multi-module systems. Analyses
of empirical results indicate that positive utility effects of
sequencing modules are outweighed by cascading errors passed on
between modules.
\end{abstract}

\section{Introduction}

Learning word pronunciation can be a hard task when the relation
between the spelling of a language and its corresponding pronunciation
is many-to-many. The English writing system and its pronunciation are
a notoriously complex example, caused by an apparent conflict between
{\em analogy}\/ and {\em inconsistency}:

\begin{description}
\item[Analogy.]  When two words or word chunks have a similar
spelling, they tend to have a similar pronunciation. This tendency
(which generalises to other language tasks as well) is usually
referred to as the {\em analogy
principle}\/\cite{saussure-course,yvon-thesis,daelemans-experience}.
\item[Inconsistency.]  Much of the analogy in English word
pronunciation is disrupted by productive and complex word
morphology, word stress, and graphematics.
\end{description}

Influential pre-Chomskyan linguistic theories have been pointing at
the analogy principle as the underlying principle for language
learning \cite{saussure-course}, and at induction as the reasoning
method for generalising from learned instances of language tasks to
new instances through analogy \cite{bloomfield-language}. However,
methods and resources (e.g., computer technology) were not available
then to demonstrate how induction through analogy could be employed to
learn and model language tasks. Partly due to this lack of
demonstrating power, Chomsky later stated 

\begin{quote}
``$\ldots$ I don't see any way of explaining the resulting final state
[of language learning] in terms of any proposed general developmental
mechanism that has been suggested by artificial intelligence,
sensorimotor mechanisms, or anything else'' (Chomsky, in
\cite{piatelli-chomsky}, p. 100).
\end{quote}

Chomsky's argument is based on the assumption that generic learning
methods such as induction cannot discover autonomously essential
levels of abstraction in language processing tasks. Applied to
morpho-phonology, the argument states that generic learning methods
are not able to discover morphology, graphematics, and stress patterns
autonomously when learning word pronunciation, although this knowledge
appears essential. Phonological and morphological theories, influenced
by Chomskyan 
theory across the board since the publication of {\sc spe}
\cite{chomsky-spe}, have generally adopted the idea of abstraction
levels in various guises (e.g., levels, tapes, tiers, grids)
\cite{goldsmith-autosegment,liberman-metrical,koskenniemi-kimmo,mohanan-phon}.
Although there is no
general consensus on which levels of abstraction can be discerned in
phonology and morphology, there is a rough, global agreement on the
fact that words can be represented on different abstraction levels as
strings of letters, graphemes, morphemes, phonemes, syllables, and
stress patterns.

According to these leading morpho-phonological theories, systems that
(learn to) convert spelled words to phonemic words in one pass, i.e.,
without making use of abstraction levels, are assumed to be unable to
generalise to new cases: going through the relevant abstraction levels
is deemed essential to yield correct conversions of previously unseen
words. This assumption implies that if one wants to build a system
that converts text to speech, one should implement explicitly the
relevant levels of abstraction. Such explicit implementations of
abstraction levels can indeed be witnessed in many state-of-the-art
speech synthesisers, implemented as (sequential) modules
\cite{allen-mitalk,daelemans-grafon}.

In this paper we challenge the assumption that levels of abstraction
must be made explicit in learning and performing the
word-pronunciation task. We do this by applying an inductive-learning
algorithm from machine learning to word pronunciation. From a wealth
of existing algorithms in machine learning
\cite{mitchell-book}, we choose {\sc igtree}
\cite{daelemans-air}, 
an inductive-learning decision-tree learning algorithm. {\sc igtree}
is a fast algorithm which has been demonstrated to be applicable to
language tasks
\cite{vandenbosch-grafon,vandenbosch-morph,daelemans-air}. We
construct {\sc igtree} decision trees for word pronunciation, and perform
empirical tests to estimate the trees' generalisation accuracy,
i.e., their ability to process new, unseen word-pronunciation
instances correctly. 

Rather than constructing and testing a single system, our approach is
to test different modularisations of the word-pronunciation task
systematically, to allow for an empirical comparison of
word-pronunciation systems with {\em and}\/ without the explicit
learning of abstraction levels. First, we train (by inductive
learning) and test a word-pronunciation model reflecting linguistic 
assumptions on abstraction levels quite closely: the model is composed
of five sequentially-coupled modules. Second, we train and test a
model in which the number of modules is reduced to three, integrating
two pairs of levels of abstraction. Third, we train and test a model
performing word pronunciation in a single pass, i.e., without modular
decomposition.

The paper is structured as follows: first, in Section~\ref{method} we
provide a description of {\sc igtree}, the data on which the {\sc
igtree} is trained and tested, and the applied experimental
methodology. Second, in Section~\ref{experiments} we introduce the
three word-pronunciation systems, and for each system we describe the
experiments performed and discuss the results obtained. In
Section~\ref{compare} we compare the three systems and analyse the
consequences of modularisation. Section~\ref{related} briefly mentions
related work on inductive learning of word
pronunciation. Section~\ref{discussion} summarises the results
obtained and lists some points of discussion.

\section{Algorithm, Data, Methodology}
\label{method}

\subsection{Algorithm: IGTREE}
\label{igtreesec}

{\sc igtree} \cite{daelemans-air} is a {\em top-down induction of
decision trees} ({\sc tdidt}) algorithm
\cite{breiman-trees,quinlan-c45}.  {\sc tdidt} is a widely-used method
in supervised machine learning \cite{mitchell-book}. {\sc
igtree} is designed as an optimised approximation of the
instance-based learning algorithm {\sc ib1-ig}
\cite{daelemans-hyphen,daelemans-air}.  In {\sc igtree}, {\em
information gain}\/ is used as a guiding function to compress a data
base of instances of a certain task into a decision tree\footnote{{\sc
igtree} can function with any feature weighting method, such as gain
ratio \cite{quinlan-c45}; for all experiments reported here,
information gain was used.}. Instances are stored in the tree as
paths of connected nodes ending in leaves which contain classification
information. Nodes are connected via arcs denoting feature
values. Information gain is used in {\sc igtree} to determine the
order in which feature values are added as arcs to the
tree. Information gain is a function from information theory, and is
used similarly in {\sc id3} \cite{quinlan-id3} and {\sc c4.5}
\cite{quinlan-c45}.

The idea behind computing the information gain of features is to
interpret the training set (i.e., the set of task instances for which
all classifications are given and which are used for training the
learning algorithm) as an information source capable of
generating a number of messages (i.e., classifications) with a certain
probability. The information entropy $H$ of such an information source
can be compared in turn for each of the features characterising the
instances (let $n$ equal the number of features), to the average
information entropy of the information source when the value of those
features are known. Data-base information entropy $H(D)$ is equal to
the number of bits of information needed to know the classification
given an instance. It is computed by equation~\ref{dbentropy}, where
$p_{i}$ (the probability of classification $i$) is estimated by its
relative frequency in the training set.
\begin{equation}
H(D) = - \sum_{i} p_{i} log_{2} p_{i}
\label{dbentropy}
\end{equation}
To determine the information gain of each of the $n$ features $f_{1}
\ldots f_{n}$, we compute the average information entropy for each
feature and subtract it from the information entropy of the data
base. To compute the information entropy for a feature
$f_{i}$, given in equation~\ref{featentropy}, we take the weighted average
information entropy of the data base restricted to each possible value
for the feature. The expression $D_{[f_{i}=v_{j}]}$ refers to those
patterns in the data base that have value $v_{j}$ for feature $f_{i}$,
$j$ is the number of possible values of $f_{i}$, and $V$ is the set of
possible values for feature $f_{i}$. Finally, $|D|$ is the number of
patterns in the (sub) data base.
\begin{equation}
H(D_{[f_{i}]}) = \sum_{v_{j} \in V} H(D_{[f_{i}=v_{j}]}) \frac{|D_{[f_{i}=v_{j}]}|}{|D|}
\label{featentropy}
\end{equation}
Information gain of feature $f_{i}$ is then obtained by
equation~\ref{funcinfogain}. 
\begin{equation}
G(f_{i}) = H(D) - H(D_{[f_{i}]})
\label{funcinfogain}
\end{equation}
In {\sc igtree}, feature-value information is stored in the decision
tree on arcs. The first feature values, stored as arcs connected to
the tree's top node, are those representing the values of the feature
with the highest information gain, followed at the second level of the
tree by the values of the feature with the second-highest information
gain, etc., until the classification information represented by a path
is unambiguous. Knowing the value of the most important feature may
already uniquely identify a classification, in which case the other
feature values of that instance need not be stored in the
tree. Alternatively, it may be necessary for disambiguation to store a
long path in the tree.

Apart from storing uniquely identified class labels at leafs, {\sc
igtree} stores at each non-terminal node information on the most
probable classification given the path so far. The most probable
classification is the most frequently occurring classification in the
subset of instances being compressed in the path being expanded.
Storing the most probable class at non-terminal nodes is essential
when processing new instances. Processing a new instance involves
traversing the tree by matching the feature values of the test
instance with arcs the tree, in the order of the feature information
gain. Traversal ends when (i) a leaf is reached or when (ii) matching
a feature value with an arc fails.  In case (i), the classification
stored at the leaf is taken as output. In case (ii), we use the most
probable classification on the last non-terminal node most recently
visited instead.

\subsection{Data Acquisition and Preprocessing}

The resource of word-pronunciation instances used in our experiments
is the {\sc celex}\index{celex@{\sc celex}} lexical data base of
English \cite{burnage-celex}. All items in the {\sc celex} data bases
contain hyphenated spelling, syllabified and stressed phonemic
transcriptions, and detailed morphological analyses.  We extracted
from the English data base of {\sc celex} all the above information,
resulting in a data base containing 77,565 unique items (word forms
with syllabified, stressed pronunciations and morphological
segmentations).

For use in experiments with learning algorithms, the data is
preprocessed to derive fixed-size instances. In the experiments
reported in this paper different morpho-phonological (sub)tasks are
investigated; for each (sub)task, an instance base (training set) is
constructed containing instances produced by {\em windowing}
\cite{sejnowski-nettalk} and attaching to each instance the
classification appropriate for the (sub)task under investigation.
Table~\ref{overall-windows-ex} displays example instances derived from
the sample word {\sf booking}. With this method, for each (sub) task
an instance base of 675,745 instances is built.

\begin{table*}
\setlength{\tabcolsep}{1mm}
\begin{center}
\begin{tabular}{||c||ccc|c|ccc||ccccc||ccc|c|ccc||cc||}
\hline
 & \multicolumn{12}{|c||}{letter-window instances} &
\multicolumn{9}{|c||}{phoneme-window instances} \\
instance & \multicolumn{3}{|c|}{left} & &
\multicolumn{3}{|c||}{right} &
\multicolumn{5}{|c||}{classifications} &
\multicolumn{3}{|c|}{left} & &
\multicolumn{3}{|c||}{right} &
\multicolumn{2}{|c||}{classif.} \\
number & \multicolumn{3}{|c|}{context} & focus &
\multicolumn{3}{|c||}{context} &
{\sc m} & {\sc a} & {\sc g} & {\sc s} & {\sc gs} &
\multicolumn{3}{|c|}{context} & focus &
\multicolumn{3}{|c||}{context} &
{\sc y} & {\sc s} \\
\hline
1  & \_ & \_ & \_ & {\sf b} & {\sf o} & {\sf o} & {\sf k}      &
1 & 1 & /b/ & 1 & /b/1 &
\_ & \_ & \_ & /b/ & /u/ & /-/ & /k/ & 1 & 1 \\
2  & \_ & \_ & {\sf b} & {\sf o} & {\sf o} & {\sf k} & {\sf i} &
0 & 1 & /u/ & 0 & /u/0 &
\_ & \_ & /b/ & /u/ & /-/ & /k/ & /\i / & 0 & 0 \\
3  & \_ & {\sf b} & {\sf o} & {\sf o} & {\sf k} & {\sf i} & {\sf n} &
0 & 0 & /-/ & 0 & /-/0 &
\_ & /b/ & /u/ & /-/ & /k/ & /\i / & /\eng / & 0 & 0 \\
4  & {\sf b} & {\sf o} & {\sf o} & {\sf k} & {\sf i} & {\sf n} & {\sf g} &
0 & 1 & /k/ & 0 & /k/0 &
/b/ & /u/ & /-/ & /k/ & /\i / & /\eng / & /-/ & 1 & 0 \\
5  & {\sf o} & {\sf o} & {\sf k} & {\sf i} & {\sf n} & {\sf g} & \_ &
1 & 1 & /\i / & 0 & /\i /0 &
/u/ & /-/ & /k/ & /\i / & /\eng / & /-/ & \_ & 0 & 0 \\
6  & {\sf o} & {\sf k} & {\sf i} & {\sf n} & {\sf g} & \_ & \_ &
0 & 1 & /\eng / & 0 & /\eng /0 &
/-/ & /k/ & /\i / & /\eng / & /-/ & \_ & \_ & 0 & 0 \\
7  & {\sf k} & {\sf i} & {\sf n} & {\sf g} & \_ & \_ & \_ &
0 & 0 & /-/ & 0 & /-/0 &
/k/ & /\i / & /\eng / & /-/ & \_ & \_ & \_ & 0 & 0 \\
\hline
\end{tabular}
\normalsize
\caption{Example of instances generated from the
word {\sf booking}, with classifications for all of the subtasks
investigated, viz. {\sc m}, {\sc a}, {\sc g}, {\sc y}, {\sc s}, and
{\sc gs}.\label{overall-windows-ex}}
\end{center}
\end{table*}
 
In the table, six classification fields are shown, one of which is a
composite field; each field refers to one of the (sub)tasks
investigated here. {\sc m} stands for morphological decomposition:
determine whether a letter is the initial letter of a morpheme (class
`1') or not (class `0'). {\sc a} is graphemic
parsing\footnote{Graphemic parsing is not represented in the {\sc
celex} data. We used an automatic alignment algorithm
\cite{daelemans-align} to determine which letters are the first or
only letters of a grapheme.}: determine whether a letter is the first
or only letter of a grapheme (class `1') or not (class `0'); a
grapheme is a cluster of one or more letters mapping to a single
phoneme. {\sc g} is grapheme-phoneme conversion: determine the
phonemic mapping of the middle letter. {\sc y} is syllabification:
determine whether the middle phoneme is syllable-initial. {\sc s} is
stress assignment: determine the stress level of the middle
phoneme. Finally, {\sc gs} is integrated grapheme-phoneme conversion
and stress assignment.  The example instances in
Table~\ref{overall-windows-ex} show that each (sub)task is phrased as
a classification task on the basis of windows of letters or phonemes
(the stress assignment task {\sc s} is investigated with both letters
and phonemes as input). Each window represents a snapshot of a part of
a word or phonemic transcription, and is labelled by the
classification associated with the middle letter of the window. For
example, the first letter-window instance {\sf \_\_\_book} is linked
with label `1' for the morphological segmentation task ({\sc m}),
since the middle letter {\sf b} is the first letter of the morpheme
{\sl book}; the other instance labelled with
morphological-segmentation class `1' is the instance with {\sf i} in
the middle, since {\sf i} is the first letter of the (inflectional)
morpheme {\sf ing}. Classifications may either be binary (`1' or `0')
for the segmentation tasks ({\sc m}, {\sc a}, and {\sc y}), or have
more values, such as 62 possible phonemes ({\sc g}) or three stress
markers (primary, secondary, or no stress, {\sc s}), or a combination
of these classes (159 combined phonemes and stress markers, {\sc gs}).

\subsection{Methodology}

Our empirical study focuses on measuring the ability of the {\sc
igtree} learning algorithm to use the knowledge accumulated during
learning for the classification of new, unseen instances of the same
(sub)task, i.e., we measure their generalisation
accuracy. \cite{weiss-learn} describe {\em $n$-fold cross validation}
($n$-fold {\sc cv}) as a procedure for measuring generalisation
accuracy. For our experiments with {\sc igtree}, we set up 10-fold
{\sc cv} experiments consisting of five steps. (i) On the basis of a
data set, $n$ partitionings are generated of the data set into one
training set containing $((n-1)/n)$th of the data set, and one test
set containing $(1/n)$th of the data set, per partitioning. For each
partitioning, the three following steps are repeated: (ii)
Information-gain values for all (seven) features are computed on the
basis of the training set (cf. Subsection~\ref{igtreesec}). (iii) {\sc
igtree} is applied to the training set, yielding an induced decision
tree (cf. Subsection~\ref{igtreesec}). (iv) The tree is tested by
letting it classify all instances in the test set, which results in a
percentage of incorrectly classified test instances. (v) When each of
the $n$ folds has produced an error percentage on test material, a
mean generalisation error of the learned model is
computed. \cite{weiss-learn} argue that by using $n$-fold {\sc cv},
preferably with $n \geq 10$, one can retrieve a good estimate of the
true generalisation error of a learning algorithm given an instance
base. Mean results can be employed further in significance
tests. In our experiments, $n = 10$, and one-tailed $t$-tests are
performed.

\section{Three word-pronunciation architectures}
\label{experiments}

\begin{figure*}
\centerline{
        \epsfxsize=12cm
        \epsfbox{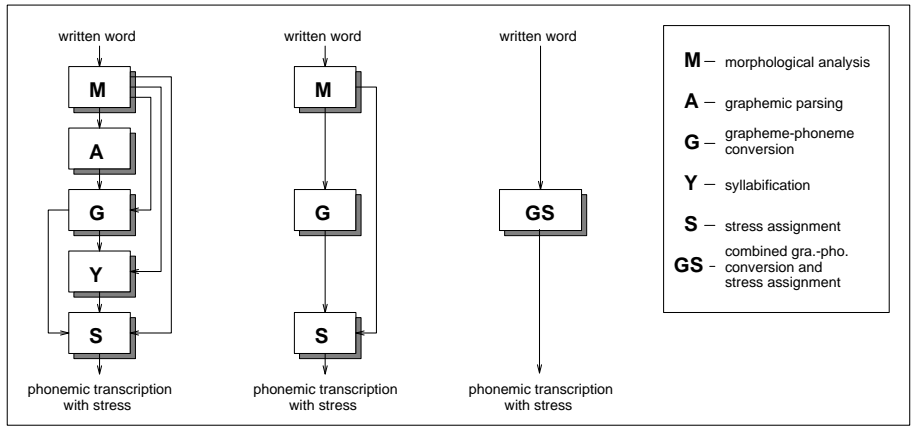}
}
\center
\caption{Architectures of the three investigated word-pronunciation
systems. Left: {\sc m-a-g-y-s}; middle: {\sc m-g-s}; right: {\sc
gs}. Rectangular boxes represent modules; the letter in the box
corresponds to the subtask as listed in the legenda (far
right). Arrows depict data flows from the raw input or a module, to a
module or the output.}
\label{threesome}
\end{figure*}

Our experiments are grouped in three series, each involving the
application of {\sc igtree} to a particular word-pronunciation
system. The architectures of these systems are displayed in
Figure~\ref{threesome}. In the following subsections, each system is
introduced, an outline is given of the experiments performed on the
system, and the results are briefly discussed.

\subsection{M-A-G-Y-S}

The architecture of the {\sc m-a-g-y-s} system is inspired by {\sc
sound1} \cite{hunnicutt-sound1,hunnicutt-tts}, the word-pronunciation
subsystem of the {\sc mitalk} text-to-speech system
\cite{allen-mitalk}. When the {\sc mitalk} system is faced with an
unknown word, {\sc sound1} produces on the basis of that word a
phonemic transcription with stress markers \cite{allen-mitalk}. This
word-pronunciation process is divided into the following five
processing components: \\

\begin{enumerate}
\setlength{\itemsep}{-0.5mm}
\item
{\em morphological segmentation}, which we implement as the module referred to as {\sc m};
\item
{\em graphemic parsing}, module {\sc a};
\item
{\em grapheme-phoneme conversion}, module {\sc g};
\item
{\em syllabification}, module {\sc y};
\item
{\em stress assignment}, module {\sc s}. \\
\end{enumerate}
 
The architecture of the {\sc m-a-g-y-s} system is visualised in the
left of Figure~\ref{threesome}.  It can be seen that the
representations include direct output from previous modules, as well
as representations from earlier modules. For example, the {\sc s}
module takes as input the syllable boundaries generated by the {\sc y}
module, but also the phoneme string generated by the {\sc g} module,
and the morpheme boundaries generated by the {\sc m} module.
 
{\sc m-a-g-y-s} is put to the test by applying {\sc igtree} in 10-fold
{\sc cv} experiments to the five subtasks, connecting the modules
after training, and measuring the combined score on correctly
classified phonemes and stress markers, which is the desired output of
the word-pronunciation system.  An individual module can be trained on
data from {\sc celex} directly as input, but this method ignores the
fact that modules in a working modular system can be expected to
generate some amount of error. When one module generates an error, the
subsequent module receives this error as input, assumes it is correct,
and may generate another error. In a five-module system, this type of
cascading errors may seriously hamper generalisation accuracy. To
counteract this potential disadvantage, modules can also be trained on
the output of previous modules. Modules cannot be expected to learn to
repair completely random, irregular errors, but whenever a previous
module makes {\em consistent} errors on a specific input, this may be
recognised by the subsequent module. Having detected a consistent
error, the subsequent module is then able to repair the error and
continue with successful processing.  Earlier experiments performed on
the tasks investigated in this paper have shown that classification
errors on test instances are indeed consistently and significantly
decreased when modules are trained on the output of previous modules
rather than on data extracted directly from {\sc celex}
\cite{vandenbosch-thesis}. Therefore, we train the {\sc m-a-g-y-s}
system, with {\sc igtree}, by training the modules of the system on
the output of predecessing modules. We henceforth refer to this type of
training as {\em adaptive}\/ training, referring to the adaptation of
a module to the errors of a predecessing module.
\begin{figure}
\centerline{
        \epsfxsize=6.5cm
        \epsfbox{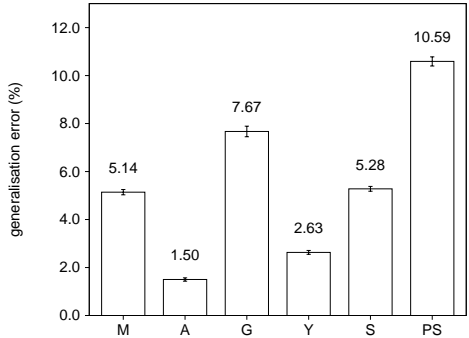}
}
\center
\caption{Generalisation errors on the {\sc m-a-g-y-s} system in
terms of the percentage of incorrectly classified test instances by
{\sc igtree} on the five subtasks {\sc m}, {\sc a}, {\sc g}, {\sc y}, and {\sc s}, and on phonemes and stress markers jointly (PS).}
\label{magys-nemlap98}
\end{figure}

Figure~\ref{magys-nemlap98} displays the results obtained with {\sc
igtree} under the adaptive variant of {\sc m-a-g-y-s}. The figure
shows all percentages (displayed above the bars; error bars on top of
the main bars indicate standard deviations) of incorrectly
classified instances for each of the five subtasks, and a joint error
on incorrectly classified phonemes with stress markers, which is the
desired output of the system. The latter classification error,
labelled PS in Figure~\ref{magys-nemlap98}, regards classification of
an instance as incorrect if either or both of the phoneme and stress
marker is incorrect.
The figure shows that the joint error on phonemes and stress markers
is 10.59\% of test instances, on average. Computed in terms of
transcribed words, only 35.89\% of all test words are converted to stressed
phonemic transcriptions flawlessly. The joint error is lower than the
sum of the errors on the {\sc g} subtask and the {\sc s} subtask,
12.95\%, suggesting that about 20\% of the incorrectly classified test
instances involve an incorrect classification of both the phoneme and
the stress marker.

\subsection{M-G-S}

The subtasks of graphemic parsing ({\sc a}) and grapheme-phoneme
conversion ({\sc g}) are clearly related. While {\sc a} attempts to
parse a letter string into graphemes, {\sc g} converts graphemes to
phonemes. Although they are performed independently in {\sc
m-a-g-y-s}, they can be integrated easily when the class-`1'-instances
of the {\sc a} task are mapped to their associated phoneme rather than
`1', and the class-`0'-instances are mapped to a {\em phonemic null},
/-/, rather than `0' (cf. Table~\ref{overall-windows-ex}).  This task
integration is also used in the {\sc nettalk} model
\cite{sejnowski-nettalk}.  A similar argument can be made for
integrating the syllabification and stress assignment modules into a
single stress-assignment module. Stress markers, in our definition of
the stress-assignment subtask, are placed solely on
the positions which are also marked as syllable boundaries (i.e., on
syllable-initial phonemes). Removing the syllabification subtask makes
finding those syllable boundaries which are relevant for stress
assignment an integrated part of stress assignment. Syllabification
({\sc y}) and stress assignment ({\sc s}) can thus be integrated in a
single stress-assignment module {\sc s}.
 
\begin{figure}
\centerline{
        \epsfxsize=6.5cm
        \epsfbox{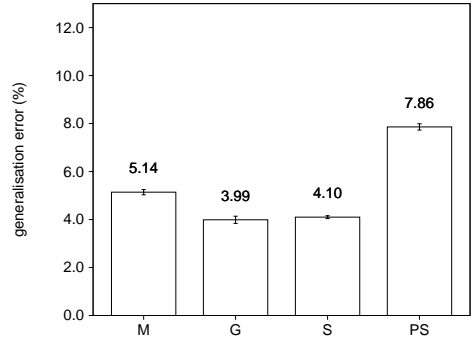}
}
\center
\caption{Generalisation errors on the {\sc m-g-s} system in
terms of the percentage of incorrectly classified test instances by
{\sc igtree} on the three subtasks {\sc m}, {\sc g}, and {\sc s}, and on phonemes and stress markers jointly (PS).}
\label{mgs-nemlap98}
\end{figure}

When both pairs of modules are reduced to single modules, the
three-module system {\sc m-g-s} is obtained.  Figure~\ref{threesome}
displays the architecture of the {\sc m-g-s} system in the middle.
Experiments on this system are performed analogous to the experiments
with the {\sc m-a-g-y-s} system; Figure~\ref{mgs-nemlap98} displays
the average percentages of generalisation errors generated by {\sc
igtree} on the three subtasks and phonemes and stress markers jointly
(the error bar labelled PS).

Removing graphemic parsing ({\sc a}) and syllabification ({\sc y}) as
explicit in-between modules yields better accuracies on the
grapheme-phoneme conversion ({\sc g}) and stress assignment ({\sc s})
subtasks than in the {\sc m-a-g-y-s} system. Both differences are
significant; for {\sc g}, ($t(19)=43.70, p<0.001$), and for {\sc s}
($t(19)=32.00, p<0.001$). The joint 
accuracy on phonemes and stress markers is also significantly
better in the {\sc m-g-s} system than in the {\sc m-a-g-y-s} system
($t(37.50, p<0.001$). Different from {\sc m-a-g-y-s}, the sum of the
errors on phonemes and stress markers, 8.09\%, is hardly more than the
joint error on PSs, 7.86\%: there is hardly an overlap in instances
with incorrectly classified phonemes and stress markers. The
percentage of flawlessly processed test words is 44.89\%, 
which is markedly better than the 35.89\% of {\sc m-a-g-y-s}.

\subsection{GS}

{\sc gs} is a single-module system in which only one classification
task is performed in one pass. The {\sc gs} task integrates
grapheme-phoneme conversion and stress assignment: to classify letter
windows as corresponding to a {\em phoneme with a stress marker}
(PS). In the {\sc gs} system, a PS can be either (i) a phoneme or a
phonemic null with stress marker `0', or (ii) a phoneme with stress
marker `1' (i.e., the first phoneme of a syllable receiving primary
stress), or (iii) a phoneme with stress marker `2' (i.e., the first
phoneme of a syllable receiving secondary stress). The simple
architecture of {\sc gs}, which does not reflect any linguistic expert
knowledge about decompositions of the word-pronunciation task, is
visualised as the rightmost architecture in Figure~\ref{threesome}. It
only assumes the presence of letters at the input, and phonemes and
stress markers at the output.  Table~\ref{overall-windows-ex} displays
example instance PS classifications generated on the basis of the word
{\sf booking}.  The phonemes with stress markers (PSs) are denoted by
composite labels. For example, the first instance in
Table~\ref{overall-windows-ex}, {\sf \_\_\_book}, maps to class label
/b/1, denoting a /b/ which is the first phoneme of a syllable
receiving primary stress.

\begin{figure}
\centerline{
        \epsfxsize=6.5cm
        \epsfbox{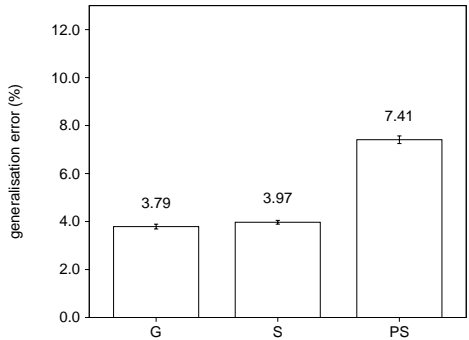}
}
\center
\caption{Percentage of generalisation errors made by {\sc igtree} on
the {\sc gs} task,  in terms of the
percentage incorrectly classified test instances as well as on
phonemes and stress assignments computed separately.}
\label{gs-nemlap98}
\end{figure}

The experiments with {\sc gs} were performed with the same data set of
word pronunciation as used with {\sc m-a-g-y-s} and {\sc m-g-s}. The
number of PS classes (i.e., all possible combinations of phonemes
and stress markers) occurring in this data base of tasks is 159.
Figure~\ref{gs-nemlap98} displays the generalisation
errors in terms of incorrectly classified test instances. The figure also
displays the percentage of classification errors made on phonemes and
stress markers computed separately.

{\sc igtree} yields significantly better generalisation accuracy on
phonemes and stress markers, both jointly and independently. In terms
of PSs, the accuracy on {\sc gs} is significantly better than that
of {\sc m-g-s} with ($t(19)=40.48, p<0.001$), and that of {\sc
m-a-g-y-s} with ($t(19)=6.90, p<0.001$). Its accuracy on flawlessly
transcribed test words, 59.38\%, is also considerably better than that
of the modular systems. Compared to accuracies reported in related
research on learning English word pronunciation 
\cite{sejnowski-nettalk,wolpert-nettalk,dietterich-id3,yvon-thesis}
and on general quality demands of text-to-speech applications, an
error of 3.79\% on phonemes and 30.62\% on words can be considered
adequate, though still not excellent
\cite{yvon-thesis,vandenbosch-thesis}.

\section{Comparisons of M-A-G-Y-S, M-G-S, and GS}
\label{compare}

%Figure~\ref{nemlap98-performances} repeats the results obtained from
%applying {\sc igtree} to each of the three systems, in terms of the
%percentages of incorrectly classified PSs.

We have given significance
results showing that, under our experimental conditions and using {\sc
igtree} as the learning algorithm, optimal generalisation accuracy on
word pronunciation is obtained with {\sc gs}, the system that does not
incorporate any explicit decomposition of the word-pronunciation task.
In this section we perform two additional comparisons of the three
systems. First, we compare the sizes of the trees constructed by {\sc
igtree} on the three systems; second, we analyse the positive and
negative effects of learning the subtasks in their specific systems'
context.
 
\subsubsection*{Tree sizes}

\begin{figure}
\centerline{
        \epsfxsize=6.5cm
        \epsfbox{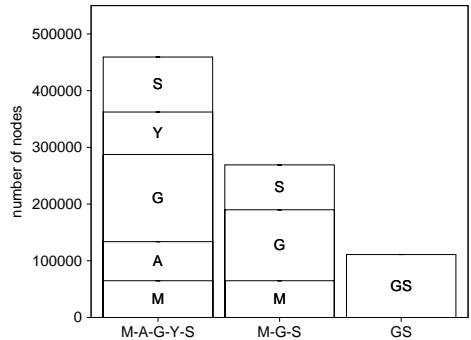} }
\center
\caption{Average numbers of nodes in the decision trees generated
by {\sc igtree} for the {\sc m-a-g-y-s}, {\sc m-g-s}, and {\sc gs} systems. Compartments indicate the numbers of nodes needed for the trees of the subtasks specified by their labels.}
\label{nemlap98-igtrees}
\end{figure}

An advantage of using less or no decompositions in terms of
computational efficiency is the total amount of memory needed for
storing the trees. Although the application of {\sc igtree} generally
results in small trees that fit well inside small computer memories
(for our modular (sub)tasks, tree sizes vary from 64,821 nodes for the
{\sc m}-modules to 153,678 nodes for the {\sc g}-module in {\sc
m-a-g-y-s}, occupying 453,747 to 1,075,746 bytes of memory), keeping
five trees in memory would not be a desirable feature for a system
optimised on memory use. Figure~\ref{nemlap98-igtrees} displays the
summed number of nodes for each of the four {\sc igtree}-trained
systems under the adaptive variant. Each bar is divided into
compartments indicating the amount of nodes in the trees generated for
each of the modular subtasks.
 
Figure~\ref{nemlap98-igtrees} shows
that the model with the best generalisation accuracy, {\sc gs}, is also the model
taking up the smallest number of nodes. The amount of nodes in
the single {\sc gs} tree, 111,062, is not only smaller than the sum of
the amount of nodes needed for the {\sc g} and {\sc s} modules in the
{\sc m-g-s} system (204,345 nodes); it is even smaller than the single
tree constructed for the {\sc g} subtask in the {\sc m-g-s} system
(125,182 nodes).

A minor difference in tree size can be seen between the trees
built for the {\sc g}-module in the {\sc m-g-s} system, 125,182 nodes,
and the {\sc g}-module in the {\sc m-a-g-y-s} system, 153,678 nodes. A
similar difference can be seen for the {\sc s}-modules, taking up
79,163 nodes in the {\sc m-g-s} system, and 96,998 nodes in the {\sc
m-a-g-y-s} system. The size of the trees built for modules appears to
increase when the module is preceded by more modules, which suggests
that {\sc igtree} is faced with a more complex task, including
potentially erroneous output from more modules, when building a tree
for a module further down a sequence of modules.

\subsubsection*{Utility effects}

The particular sequence of the five modules as in the {\sc m-a-g-y-s}
system reflects a number of assumptions on the {\em utility}\/ of
using output from one subtask as input to another
subtask. Morphological knowledge is useful as input to
grapheme-phoneme conversion (e.g., to avoid pronouncing {\sf ph} in
{\sf loophole} as /f/, or {\sf red} in {\sf barred} as /r\niepsilon
d/); graphemic parsing is useful as input to grapheme-phoneme
conversion (e.g., to avoid the pronunciation of {\sf gh} in {\sf
through}); etc. Thus, feeding the output of a module $A$ into a
subsequent module $B$ implies that one expects to perform better on
module $B$ with $A$'s input than without. The accuracy results
obtained with the modules of the {\sc m-a-g-y-s}, {\sc m-g-s}, and {\sc
gs} systems can serve as tests for their respective underlying utility
assumptions, when they are compared to the accuracies obtained with
their subtasks learned in isolation.

To measure the {\em utility effects}\/ of including the outputs of
modules as inputs to other modules, we performed the following
experiments: \\

\begin{enumerate}
\item
We applied {\sc igtree} in 10-fold {\sc cv} experiments to each of the
five subtasks {\sc m}, {\sc a}, {\sc g}, {\sc y}, and {\sc s}, only
using
letters (with the {\sc m}, {\sc a}, {\sc g}, and {\sc s}
subtasks) or phonemes (with the {\sc y} and the {\sc s} subtasks) as
input, and their respective classification as output
(cf. Table~\ref{overall-windows-ex}). The input is directly extracted
from {\sc celex}. These
experiments provide the baseline score for 
each subtask, and are referred to as the {\em isolated}\/ experiments.
\item
We applied {\sc igtree} in 10-fold {\sc cv} experiments to all
subtasks of the {\sc m-a-g-y-s}, {\sc m-g-s}, and {\sc gs} systems,
training {\em and}\/ testing on input extracted directly from {\sc
celex}. The results from these experiments reflect what would be the
accuracy of the modular systems when each module would perform
perfectly flawless. We refer to these experiments as {\em ideal}. \\
\end{enumerate}

With the results of these experiments we measure, for each subtask in
each of the three systems, the utility effect of including the input
of preceding modules, for the ideal case (with input straight from
{\sc celex}) as well as for the actual case (with input from preceding
modules). A utility effect is the difference between {\sc igtree}'s
generalisation error on the subtask in modular context (either ideal
or actual) and its accuracy on the same subtask in
isolation. Table~\ref{utility-overview} lists all computed utility
effects.

\begin{table}
\setlength{\tabcolsep}{1.7mm}
\begin{center}
\begin{tabular}{||c||r|rr|rr||}
\hline
sub- & \multicolumn{5}{c||}{\% generalisation error} \\
task & {\scriptsize isolated} & {\scriptsize ideal} & {\scriptsize (utility)} & {\scriptsize actual} & {\scriptsize (utility)} \\
\hline
\hline
\multicolumn{6}{|c|}{\sc m-a-g-y-s} \\
\hline
{\sc m} & 5.14 &  5.14 &    (0.00) &  5.14 &    (0.00) \\
{\sc a} & 1.39 &  1.66 & ($-$0.27) &  1.50 & ($-$0.11) \\
{\sc g} & 3.72 &  3.68 & ($+$0.04) &  7.67 & ($-$3.95) \\
{\sc y} & 0.45 &  0.75 & ($-$0.30) &  2.63 & ($-$2.16) \\
{\sc s} & 7.96 &  2.67 & ($+$5.29) &  5.28 & ($+$2.68) \\
\hline
\multicolumn{6}{|c|}{\sc m-g-s} \\
\hline
{\sc m} & 5.14 &  5.14 &    (0.00) &  5.14 &    (0.00) \\
{\sc g} & 3.72 &  3.66 & ($+$0.06) &  3.99 & ($-$0.27) \\
{\sc s} & 7.96 &  3.97 & ($+$3.99) &  4.10 & ($+$3.86) \\
\hline
\multicolumn{6}{|c|}{\sc gs} \\
\hline
{\sc g} & 3.72 &    --  &   --  &  3.79 & ($-$0.07) \\
{\sc s} & 4.71 &    --  &   --  &  3.97 & ($+$0.74) \\
\hline
\end{tabular}
\caption{Overview of utility effects of learning subtasks ({\sc m},
{\sc a}, {\sc g}, {\sc y}, and {\sc s}) as modules or partial tasks in
the {\sc m-a-g-y-s}, {\sc m-g-s}, and {\sc gs} systems. For each
module, in each system, the utility of training the module with
ideal data (middle) and actual, modular data under the adaptive
variant (right), is compared against the accuracy obtained with
learning the subtasks in isolation (left). Accuracies are given in
percentage of incorrectly classified test
instances.\label{utility-overview}}
\end{center}
\end{table}
 
For the case of the {\sc m-a-g-y-s} system, it can be seen that the
only large utility effect, even in the ideal case, could be obtained
with the stress-assignment subtask. In the isolated case, the input
consists of phonemes; in the {\sc m-a-g-y-s} system, the input
contains morpheme boundaries, phonemes, and syllable boundaries. The
ideal positive effect on the {\sc s} module of 5.29\% less errors
turns out to be a positive effect of 2.68\% in the actual
system. The latter positive effect is outweighed by a rather large
negative utility effect on the grapheme-phoneme conversion task of
$-3.95$\%. Both the {\sc a} and {\sc y} subtasks do not profit from
morphological boundaries as input, even in the ideal case; in the
actual {\sc m-a-g-y-s} system, the utility effect of including
morphological boundaries from {\sc m} and phonemes from {\sc g} in the
syllabification module {\sc y} is markedly negative: $-2.16$\%.

In the {\sc m-g-s} system, the utility effects are generally less
negative than in the {\sc m-a-g-y-s} system. There is a small utility
effect in the ideal case with including morphological boundaries as
input to grapheme-phoneme conversion; in the actual {\sc m-g-s}
system, the utility effect is negative ($-0.27$\%). The
stress-assignment module benefits from including morphological
boundaries and phonemes in its input, both in the ideal case and in
the actual {\sc m-g-s} system. 

The {\sc gs} system does not contain separate modules, but it is
possible to compare the errors made on phonemes and stress assignments
separately to the results obtained on the subtasks learned in
isolation. Grapheme-phoneme conversion is learned with almost the same
accuracy when learned in isolation as when learned as partial task of
the {\sc gs} task. Learning the grapheme-phoneme task, {\sc igtree} is
neither helped nor hampered significantly by learning stress assignment
simultaneously. There is a positive utility effect in learning stress
assignment, however. When stress assignment is learned in isolation
with letters as input, {\sc igtree} classifies 4.71\% of test
instances incorrectly, on average. (This is a lower error than obtained
with learning stress assignment on the basis of phonemes, indicating
that stress assignment should take letters as input rather than
phonemes.) When the stress-assignment task is learned along
with grapheme-phoneme conversion in the {\sc gs} system, a marked
improvement is obtained: 0.74\% less classification errors are made. 

Summarising, comparing the accuracies on modular subtasks to the
accuracies on their isolated counterpart tasks shows only a few
positive utility effects in the actual system, all obtained with
stress assignment. The largest utility effect is found on the
stress-assignment subtask of {\sc m-g-s}. However, this positive
utility effect does not lead to optimal accuracy on the {\sc s}
subtask; in the {\sc gs} system, stress assignment is performed with
letters as input, yielding the best accuracy on stress assignment
in our investigations, viz. 3.97\% incorrectly classified test instances.

\section{Related work}
\label{related}

The classical {\sc nettalk} paper by \cite{sejnowski-nettalk} can be
seen as a primary source of inspiration for the present study; it has
been so for a considerable amount of related work. Although it has
been criticised for being vague and presumptuous and for presenting
generalisation accuracies that can be improved easily with other
learning methods
\cite{stanfill-nettalk,wolpert-nettalk,weijters-nettalk2,yvon-thesis},
it was the first paper to investigate grapheme-phoneme conversion as
an interesting application for general-purpose learning algorithms.
However, few reports have been made on the joint accuracies on stress
markers and phonemes in work on the {\sc nettalk} data.  To our
knowledge, only \cite{shavlik-nettalk} and \cite{dietterich-id3}
provides such reports. In terms of incorrectly processed test
instances, \cite{shavlik-nettalk} obtain better performance with the
back-propagation algorithm trained on distributed output (27.7\%
errors) than with the {\sc id3} \cite{quinlan-id3} decision-tree
algorithm (34.7\% errors), both trained and tested on small
non-overlapping sets of about 1,000 instances. \cite{dietterich-id3}
reports similar errors on similarly-sized training and test sets
(29.1\% for {\sc bp} and 34.4\% for {\sc id3});
with a larger training set of 19,003 words from the {\sc nettalk} data and an
input encoding fifteen letters, previous phoneme and stress
classifications, some domain-specific features, and error-correcting
output codes {\sc id3} generates 8.6\% errors on test instances
\cite{dietterich-id3}, which does not compare favourably to the results
obtained with the {\sc nettalk}-like {\sc gs} task (a valid comparison cannot
be made; the data employed in the current study contains considerably
more instances).

An interesting counterargument against the representation of the
word-pronunciation task using fixed-size windows, put forward by
Yvon \cite{yvon-thesis}, is that an inductive-learning approach to
grapheme-phoneme conversion should be based on associating
variable-length chunks of letters to variable-length chunks of
phonemes. The chunk-based approach is shown to be
applicable, with adequate accuracy, to several corpora, including
corpora of French word pronunciations and, as mentioned above, the
{\sc nettalk} data \cite{yvon-thesis}. Experiments on other (larger)
corpora, comparing both approaches, would be needed to analyse their
differences empirically.

\section{Discussion}
\label{discussion}

We have demonstrated that a decision-tree learning algorithm, {\sc
igtree}, is able to learn English word pronunciation with modest to
adequate generalisation accuracy: the less the learning task is
decomposed in subtasks, the more adequate the generalisation
accuracy obtained by {\sc igtree} is. The best generalisation
accuracy is obtained with the {\sc gs} system, which does not
decompose the task at all. The general disadvantage of the
investigated modular systems is that modules do not perform their
tasks flawlessly, while their expert-based decompositions do assume
flawless performance. In practice, modules produce a considerable amount of
irregular errors which cause subsequent modules to generate subsequent
`cascading' errors. Only the subtask of stress assignment is shown to
be learned more successfully on the basis of modular input.

The best-performing system, {\sc gs}, is trained to map windows of
letters to combined class labels representing phonemes and stress
markers. Compared to the {\sc m-a-g-y-s} and
{\sc m-g-s} systems, the {\sc gs} system (i) lacks an explicit 
morphological segmentation and (ii) learns stress
assignment jointly with grapheme-phoneme conversion on the basis of
letter windows rather than phoneme windows. These two advantageous
properties of the {\sc gs} system lead to three suggestions. First, it
appears better to leave morphological segmentation an implicit
subtask; it can be left to the learning algorithm to extract the necessary
morphological information needed to disambiguate between alternative
pronunciations directly from the letter-window input. Second,
letter-window instances provide the most reliable source of input for
both grapheme-phoneme conversion and stress assignment. Third, stress
assignment and grapheme-phoneme conversion can be integrated in one
task, i.e., to map letter instances to `stressed phonemes'.

A warning on the scope of these suggestions needs to be issued. The
results described here are not only dependent of the resource ({\sc
celex}) and the (sub)task definitions (classification of windowed
instances), but also on the use of {\sc igtree} as the learning
algorithm. The {\sc celex} data appears robust and provides an
abundance of English word pronunciations, not an inappropriately
skewed subset of the English vocabulary. The windowing method appears
a salient method to rephrase language tasks as classification tasks
based on fixed-length inputs. It is not clear, however, to what extent
{\sc igtree} can be held responsible for the low accuracy on {\sc
m-a-g-y-s} and {\sc m-g-s}; {\sc igtree} may be negatively sensitive
in terms of generalisation accuracy to irregular errors in the
input of a modular subtask. Although irregular errors are an inherent
problem for modular systems, other learning algorithms may be able to
handle such errors differently. Experiments with back-propagation
learning applied to the same modular systems show
siginficantly worse performance than 
that of {\sc igtree} \cite{vandenbosch-thesis}. It might be possible
that instance-based learning algorithms 
(e.g., {\sc ib1-ig} \cite{daelemans-hyphen,daelemans-air}), which have
been demonstrated to outperform {\sc igtree} on several language tasks
\cite{daelemans-cl,vandenbosch-morph,vandenbosch-thesis}, perform
better on the modular systems.  Although such systems trained with {\sc ib1-ig} would be computationally
rather inefficient \cite{vandenbosch-thesis}, employing {\sc ib1-ig}
in learning modular subtasks may lead to other differences in
accuracy between modular systems.

A conclusion to be drawn from our study is that it is possible
to learn the complex language task of English word pronunciation with
a general-purpose inductive-learning algorithm, with an adequate level
of generalisation accuracy. The results suggest that the necessity
of decomposing word-pronunciation in several subtasks should be
reconsidered carefully when designing an accuracy-oriented
word-pronunciation system. Undesired errors generated by sequenced
modules may outweigh the desired positive utility effects easily.  

\subsection*{Acknowledgements}
We thank Eric Postma, Maria Wolters, David Aha, Bertjan Busser, Jakub
Zavrel, and the other members of the Tilburg ILK group for fruitful
discussions.

\end{document}